\pdfoutput=1
\documentclass{INTERSPEECH2023}
\usepackage{multirow}
\usepackage{booktabs}
\usepackage{xcolor}
\usepackage{newtxmath}
\usepackage{makecell}
\usepackage{CJKutf8}
\usepackage{subcaption}
\usepackage{url}
\usepackage{color}
\usepackage{colortbl}
\usepackage{makecell}
\usepackage{graphicx}
\usepackage{grffile}

\interspeechcameraready

\title{Accurate and Structured Pruning for Efficient Automatic Speech Recognition}

\name{Huiqiang Jiang$^1$, Li Lyna Zhang$^1$, Yuang Li$^2$, Yu Wu$^1$, Shijie Cao$^1$,\\ Ting Cao$^1$, Yuqing Yang$^1$, Jinyu Li$^3$, Mao Yang$^1$, Lili Qiu$^1$}
\address{$^1$Microsoft Research, $^2$University of Cambridge, $^3$Microsoft Azure Speech}
\email{\{hjiang, lzhani, wu.yu, shijiecao, ting.cao, yuqing.yang, jinyli, maoyang, liliqiu\}@microsoft.com, yl807@eng.cam.ac.uk}

\begin{document}

\maketitle

\begin{abstract}
    Automatic Speech Recognition (ASR) has seen remarkable advancements with deep neural networks, such as Transformer and Conformer. However, these models typically have large model sizes and high inference costs,
posing a challenge to deploy on resource-limited devices.
In this paper, we propose a novel compression strategy that leverages structured pruning and knowledge distillation to reduce the model size and inference cost of the Conformer model while preserving high recognition performance. Our approach utilizes a set of binary masks to indicate whether to retain or prune each Conformer module,
and employs $L_0$ regularization to learn the optimal mask values. To further enhance pruning performance,  we use a layerwise distillation strategy to transfer knowledge from unpruned to pruned models. Our method outperforms all pruning baselines on the widely used LibriSpeech benchmark, achieving a 50\% reduction in model size and a 28\% reduction in inference cost with minimal performance loss.
\end{abstract}

\noindent\textbf{Index Terms}: 
ASR, Model Compression, Structured Pruning

\section{Introduction}

Large models such as Conformer~\cite{gulati2020conformer}, wav2vec 2.0~\cite{baevski2020wav2vec}, and Whisper~\cite{radford2022robust} have achieved remarkable success in Automatic Speech Recognition (ASR) systems, yet they come at a high cost in terms of storage, memory, and inference latency.
For instance, the Whisper large model comprises 1.5 billion parameters, making it extremely difficult to deploy on resource-constrained scenarios such as client edge devices. Therefore, it is crucial to compress ASR models for practical deployment.

Knowledge distillation~\cite{li2014learning, hinton2015distilling} have been demonstrated to be effective in producing a faster model for ASR~\cite{huang2023improving,xu2022self,rathod2022multi,wei2022model}. However, these methods have several limitations. First, it necessitates a meticulous design to customize the high-quality smaller model~\cite{chang2022distilhubert,meng2023compressing}, which requires expertise and can be expensive, given the various deployment requirements of real-world ASR applications. Second, distillation involves a full training from scratch process~\cite{wang2022exploring,lee2022fithubert}, which is time-consuming.

Weight pruning can efficiently produce a small model that meets a given resource constraint. In particular, magnitude pruning~\cite{han2015deep,han2015learning}, which preserves model parameters with high absolute values, is the most widely used method for pruning an ASR model~\cite{lai2021parp, kim21m_interspeech, prasad2022pada}. However, it can often lead to unsatisfactory performance under a high compression ratio. Moreover, while these weight pruning methods enable the removal of weights at arbitrary locations, resulting in sparsity, these sparse models are usually unstructured, which hinders actual efficiency benefits. This is because running them in standard hardware usually requires reconstructing the original dense shape, leading to little improvement in latency~\cite{lai2021parp,nn_pruning}.

In this work, we propose a novel structured pruning approach to directly speed up ASR inference of a Conformer-Transducer~\cite{gulati2020conformer} model, while preserving high accuracy. Our approach leverages learnable binary masks to determine whether to retain or prune distinct components of the Conformer encoder within a hybrid pruning granularity, including: 
\textit{(i)} an entire attention head in a multi-head self-attention layer, \textit{(ii)} a specific dimension in the intermediate layer of the feed-forward network (FFN), \textit{(iii)} the entire convolution module in each Conformer block, and \textit{(iv)} the hidden dimension. To determine the values of the masks, we employ an augmented $L_0$ regularization approach that makes the discrete masks differentiable. In contrast to magnitude pruning, our approach jointly learns the masks and updates model parameters to achieve optimal pruning decisions. Furthermore, we introduce a layerwise knowledge distillation technique to further enhance the pruning decisions and performance by transferring layerwise hidden states from the unpruned model to the pruned model.

We evaluate our method  on the widely-used LibriSpeech benchmark~\cite{panayotov2015librispeech} under various compression ratios. Experimental results show that our method can achieve up to 50\% parameter pruning with minimal degradation in performance for Conformer-Transducer. Furthermore, the pruned model reduces the real-time factor by 28\% on a CPU device.

\section{Method}
\begin{figure*}[t]
  \centering
\resizebox{1.75\columnwidth}{!}{
     \begin{subfigure}[t]{0.26\linewidth}
         \centering
         \includegraphics[width=\linewidth]{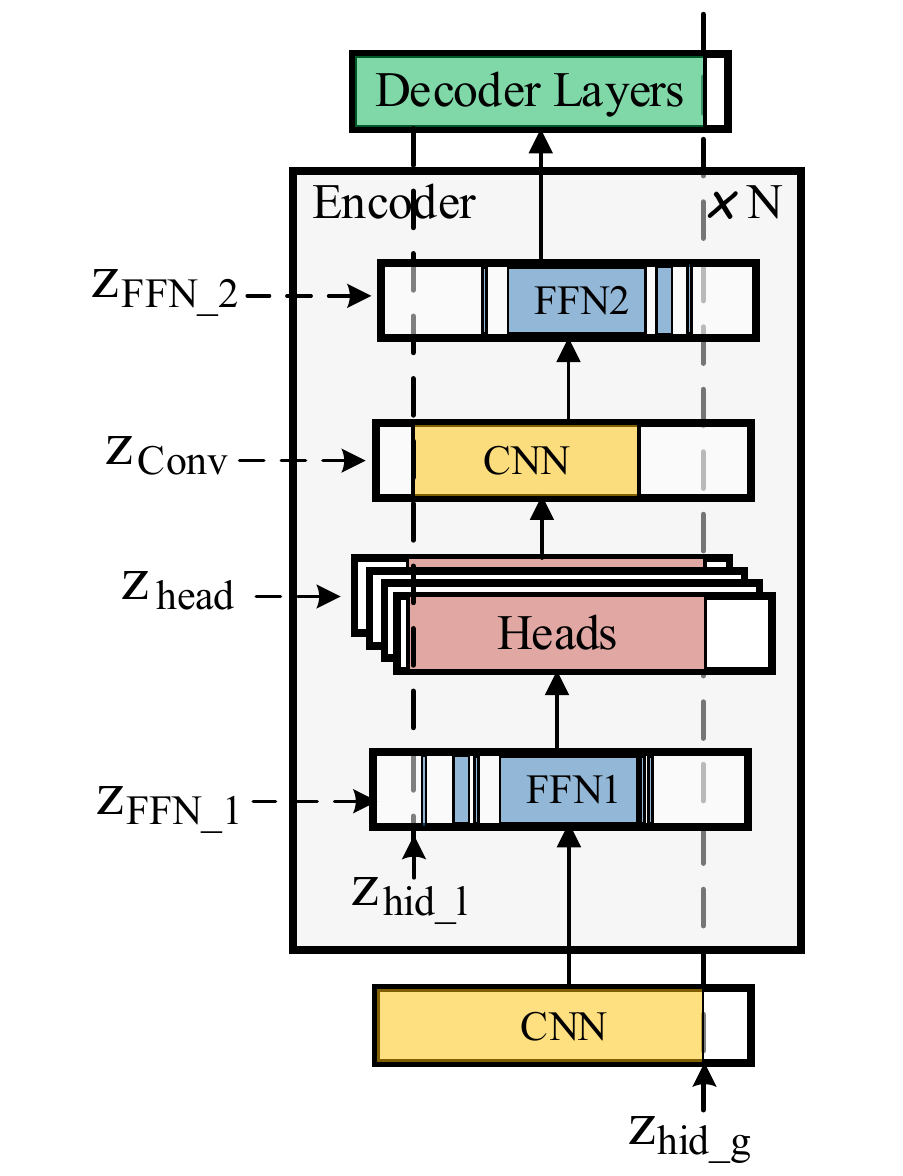}
         \caption{Structured pruning units}
         \label{subfig:pruning_space}
     \end{subfigure} 
    \begin{subfigure}[t]{0.65\linewidth}
         \centering
         \includegraphics[width=\linewidth]{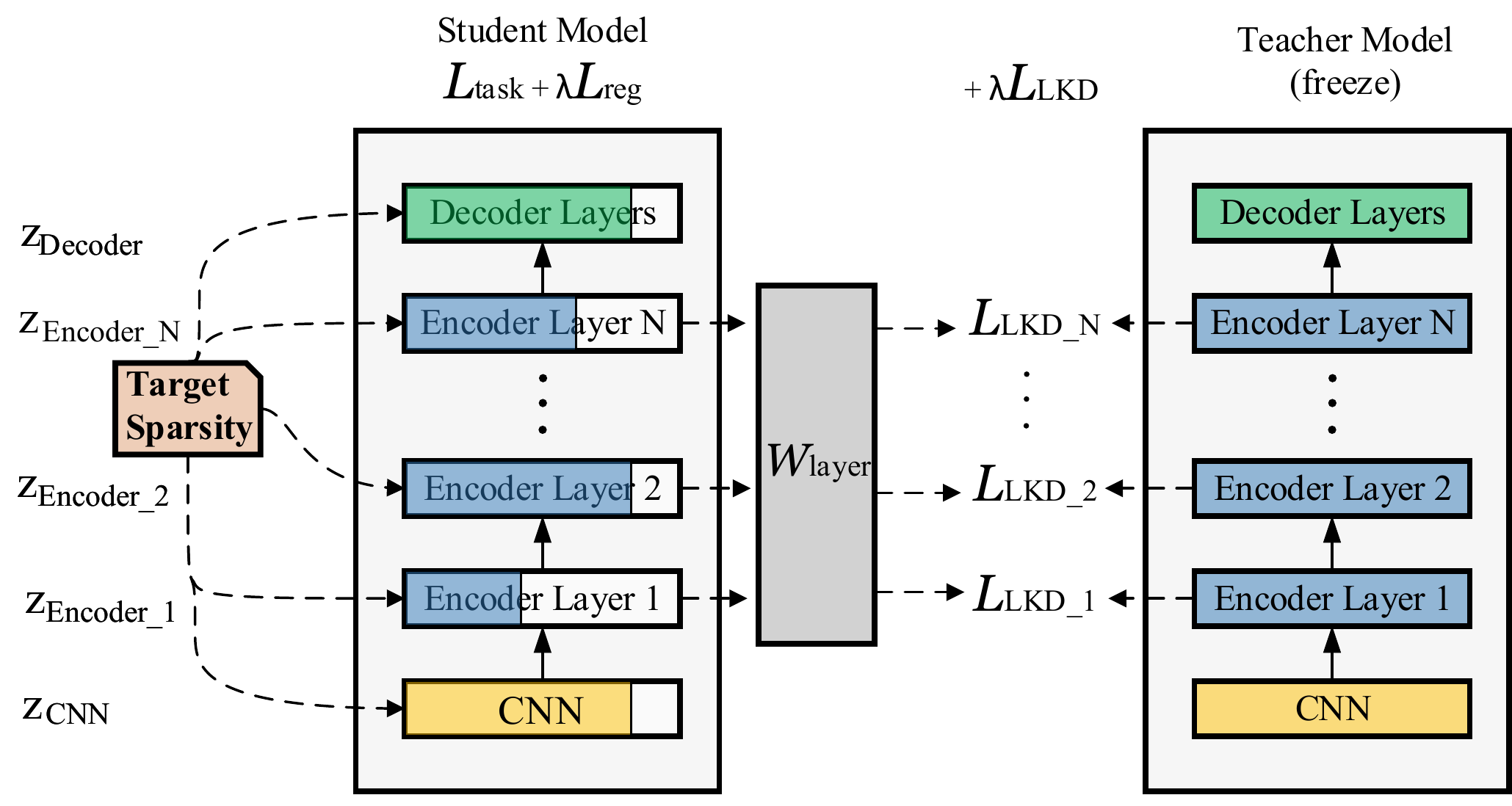}
         \caption{A visualization of the pruning pipeline}
         \label{subfig:pipeline}
     \end{subfigure}
}
 \vspace{-2ex}
  \caption{ (a) the hybrid pruning granularity designed for Conformer model; (b) The overview of our pruning algorithm. By combining knowledge distillation and pruning, our approach learns an optimal set of pruning masks.  $W_{\text{layer}}$ is the spatial transformation matrix in layer-wise knowledge distillation (Eq.~\ref{eq:finalloss}). %The decoder layers represent the prediction network and joint network modules in the Conformer-Transducer.
 }
  \label{fig:frameworks}
\end{figure*}

In this section, we present our approach for expediting ASR inference on hardware resource-limited scenarios.
We use Conformer~\cite{gulati2020conformer} as the representative model due to its wide usage in ASR. Our approach consists of two steps. First, we introduce a hybrid structured pruning strategy tailored to the CNN-Transformer architecture of the Conformer model, which assigns binary masks to various modules such as attention heads and hidden dimensions to indicate their removal or retention. 
Second, we leverage the power of knowledge distillation and incorporate an $L_0$ regularization pruning algorithm to jointly optimize the binary pruning masks and the model parameters in order to strike the best balance between speed and accuracy.

\subsection{Hybrid pruning granularity}
\label{subsec:hybrid_pruning_space}

The model pruning problem can be formulated as:

\begin{equation}
\begin{aligned}
\min_{\boldsymbol{\theta}, \mathbf{z}} \mathbb{E}_{\mathbf{z}}\left[\frac{1}{N} \sum_{i=1}^N \mathcal{L}_{\text{task}}\left(\mathbf{x}_i, \mathbf{y}_i ; \boldsymbol{\theta} \odot \mathbf{z}\right) + \lambda\lVert \mathbf{z}\rVert_0\right] \\
\label{eq:problem}
\end{aligned}
\end{equation}
where $\theta$ denotes the model weights, $\mathbf{z}\in$\{0,1\} is the binary mask introduced by pruning, where 0 indicates removal and 1 indicates retention of a corresponding parameter. $N$, $\mathbf{x}_i$, $\mathbf{y}_i$ and $\mathcal{L}_{\text{task}}$ are the sample size, input, labels, and task loss of ASR task. $\lambda$ is the hyperparameter that controls the tradeoff between accuracy and sparsity.  $\|\cdot\|_0$ denotes the 0-norm loss coefficient.

Previous research~\cite{lai2021parp,kim21m_interspeech} has mostly employed magnitude-based unstructured pruning, which allows the pruning of unimportant parameters at arbitrary locations based on weights value. However, this method often yields poor performance with higher levels of sparsity and does not directly offer acceleration benefits. Recently, unstructured~\cite{movement, molchanov2019importance} and structured~\cite{lagunas-etal-2021-block, xia-etal-2022-structured,swiftpruner} pruning methods have been proposed in other domains. However, these methods are not specifically designed for ASR tasks and may result in suboptimal performance. %, but they faced difficulties in direct application to the ASR.

To address this, our objective is to structurally remove unimportant parameters for ASR models, thereby enabling faster inference on standard hardware while preserving accuracy. 
While prior research has primarily focused on pruning transformer or CNN models, our approach is tailored to the Conformer architecture, which features a unique combination of convolution and transformer modules. To achieve optimal results, we must develop a specific pruning policy that accounts for all four modules that comprise each Conformer encoder layer: a feed-forward module (FFN$_1$), a multi-head self-attention module (MHA), a convolution module, and a second feed-forward module (FFN$_2$).
At the same time, we also prune the embedding layers and the decoding layers.
To attain this goal, we propose four types of binary masks $\mathbf{z}\in$\{0,1\}  to control the sparsity of different modules, as illustrated in Fig.~\ref{subfig:pruning_space}.

\begin{itemize}
	\item \textsc{Head mask $\mathbf{z}_{\text{head}}$}. We use $\mathbf{z}_{\text{head}_j}^i$ to determine whether $j$-th head of the $i$-th encoder layer should be kept or removed.
	\item \textsc{FFN intermediate mask $\mathbf{z}_{\text{FFN}}$ }. We use masks $\mathbf{z}_{\text{FFN}_j}^i$ to prune the $i$-th FFN $j$-th channel in intermediate dimensions. %Specifically, we mask both FFN modules in a Conformer encoder.
	\item \textsc{Conv module mask $\mathbf{z}_{\text{conv}}$}. Although the convolution module has comparatively few parameters, it contributes significantly to the total number of floating-point operations (FLOPs) in the model. To maximize inference efficiency, we introduce a gate mask, denoted as $\mathbf{z}_{\text{conv}}^i$, which allows for the pruning of the $i$-th entire convolution module in each block.
	\item \textsc{Layer-wise hidden mask $\mathbf{z}_{\text{hidden}}$}. We prune the hidden dimensions of the encoder layers to enable more flexibility 
 by using global and local masks, denoted as $\mathbf{z}_{\text{hidden}}$. The global mask, $\mathbf{z}_{\text{hid-g}}$, determines the size of hidden representations across all encoder layers, while the local mask, $\mathbf{z}_{\text{hid-l}}$, controls the hidden size within an encoder layer.
\end{itemize}

\subsection{Learning the optimal mask under a desired sparsity}
\label{sec:prune}
Based on the hybrid pruning granularity proposed in Section~\ref{subsec:hybrid_pruning_space}, this paper designs an $L_0$-based pruning method (as shown in Fig.~\ref{subfig:pipeline}) that gradually reaches a target sparsity  during the pruning training process and learns the optimal sparsity allocation according to the ASR task loss. Moreover, we combine a layer-wise knowledge distillation with pruning to further enhance the performance.

\noindent\textbf{Sparsity-aware Constraint}. With the use of pruning masks, we can measure the number of retained model parameters and, thus, calculate the inference cost. Formally, the expected model parameters after pruning can be calculated as follows:
\begin{equation}
	\begin{aligned}
	\text{S}(\mathbf{z}) = \sum_{i=1}^{L} [\text{S}(\mathbf{z})_{\text{FFN\_1}}^i + \text{S}(\mathbf{z})_{\text{Attn}}^i + \text{S}(\mathbf{z})_{\text{CNN}}^i + \text{S}(\mathbf{z})_{\text{FFN\_2}}^i]
	\end{aligned}
\label{eq:model_size_score}
\end{equation}
where $L$ is the number of encoder layers and $\text{S}(\mathbf{z})^i$ denote the parameters of different modules in $i$-th layer after pruning. We calculate the retained parameters of each module as  follows:
\begin{equation}
\begin{aligned}
    &\text{S}(\mathbf{z})_{\text{FFN\_1}}^i = \|\mathbf{z}_{\text{FFN\_1}}^i\|_0 \times(2\times\|\mathbf{z}_{i,\text{hid-g}}\|_0  + \|\mathbf{z}_{i,\text{hid-l}}\|_0) \\
    &\text{S}(\mathbf{z})_{\text{Attn}}^i = \|\mathbf{z}_{\text{hid-l}}^i\|_0 \times \|\mathbf{z}_{\text{head}}^i\|_0 \times \mathbb{S}_{\text{head}} \\
    &\text{S}(\mathbf{z})_{\text{CNN}}^i = \|\mathbf{z}_{\text{hid-l}}^i\|_0
		\times \|\mathbf{z}_{\text{conv}}^i\|_0 \times \mathbb{S}_{\text{conv}}  \\
    &\text{S}(\mathbf{z})_{\text{FFN\_2}}^i = \|\mathbf{z}_{\text{FFN\_2}}^i\|_0 \times(2\times\|\mathbf{z}_{i,\text{hid-l}}\|_0  + \|\mathbf{z}_{i,\text{hid-g}}\|_0)
    \end{aligned}
\label{eq:module_size}
\end{equation}

Similarly, the FLOPs value $\text{S}(\mathbf{z})_{\text{FLOPs}}$ of the pruned model can be obtained, thus providing an estimation of the computational cost that would be required  on edge devices.

\noindent\textbf{Sparse Allocation Optimization based on $L_0$ Regularization}.
Now, we introduce a method for determining the values of masks for pruning. Unlike magnitude pruning, which sets masks based on the values of weights, our proposed approach seeks to jointly learn masks and model parameters so that the resulting model can achieve minimal task empirical risk while meeting the desired sparsity. To achieve this, a $L_0$ regularization term $\mathcal{L}_{\text{reg}}$~\cite{louizos2017learning} is added to the optimization problem, turning the pruning problem into an end-to-end learning problem.
\begin{equation}
	\label{eq:loss1}
	\begin{aligned}
		\mathcal{L}=\mathcal{L}_{\text{task}}(\boldsymbol{\theta},\mathbf{z})+\lambda \mathcal{L}_{\text{reg}} (\mathbf{z})
	\end{aligned} 
\end{equation}
However, these binary masks $\mathbf{z}$ are discrete values and non-differentiable~\cite{bengio2013estimating}. To address this challenge, we use the reparameterization method with the hard concrete distribution proposed by~\cite{louizos2017learning}. Specifically, masks $\mathbf{z}$ can be sampled from a random variable $\mathbf{u} \sim U(0,1)$ distribution:
\begin{equation}
	\begin{aligned}
		\mathbf{t} &=
  \operatorname{sigmoid}(\log \frac{\mathbf{u}}{1-\mathbf{u}}+\boldsymbol{\alpha});
		\mathbf{z} =\min (1, \max (0, \mathbf{t} \times(r-l)+l))
		\label{eq:hard_concrete}
	\end{aligned}
\end{equation}
Where $l < 0$ and $r > 0$ are two constants, with the common practice of setting $l$ to -0.1 and $r$ to 1.1, parameters $\boldsymbol{\alpha}$ are learnable. 
Then, the $L_0$ regularization term can be regulated as a function of the cumulative density function $Q(\mathbf{t}\leq 0|\boldsymbol{\alpha})$, which will be differentiable. The expectation of mask $\mathbf{z}$  in Eq.~\ref{eq:problem} can be expressed as,
\begin{equation}
\begin{aligned}
\mathbb{E}_{\mathbf{z}}\|\mathbf{z}\|_0 = 1 - Q(\mathbf{t}\leq 0|\boldsymbol{\alpha}) = 1 - \operatorname{sigmoid}(\log \frac{\mathbf{t} - l}{r- \mathbf{t}}-\boldsymbol{\alpha})
\label{eq:expect_l0}
\end{aligned}
\end{equation}

Next, we augment $L_0$ regularization to better control the achieved model sparsity. The $\lambda$ in Eq.~\ref{eq:loss1} is used to balance the trade-off between model accuracy and sparsity. 
However, it requires careful hyper-parameter tuning to make sure it converges to a desired sparsity. To effectively control the final model sparsity, we follow~\cite{lagrangian} and replace the original $L_0$ regularization with a Lagrangian multiplier. 
Let $s$ denote the target model size after pruning, and $\text{S}(\mathbf{z})$  denote the expected model parameters determined by the masks $\mathbf{z}$ in Eq.~\ref{eq:model_size_score}. We impose an equality constraint $\text{S}(\mathbf{z})$ =$s$ by introducing a violation penalty:
\begin{equation}
	\label{eq:loss3}
	\begin{aligned}
    \mathcal{L}_{reg}(\mathbf{z})=\lambda_1(\text{S}(\mathbf{z}) - s) + \lambda_2 (\text{S}(\mathbf{z})- s)^2
	\end{aligned} 
\end{equation}
where $\lambda_1$ and $\lambda_2$ are automatically adjusted using the  AdamW optimizer~\cite{loshchilov2017decoupled}.

\noindent\textbf{Combing Pruning and Knowledge Distillation (KD)}.
The combination of knowledge distillation and pruning has been shown to perform better than pruning alone~\cite{movement,nn_pruning}. Previous works typically adopt a cross-entropy distillation loss between the teacher and student models in the final prediction layer. Our method, however, employs a layer-by-layer distillation to best utilize the layerwise semantic information in the teacher model. Through this guidance, the pruned model is encouraged to preserve more detailed layerwise semantic knowledge, leading to improved accuracy.

Specifically, we use the original unpruned model as the teacher and the model under pruning as the student. For a Conformer model with $L$ encoder layers, our layerwise knowledge distillation aims to minimize the average distance between the hidden states of each encoder layer of the teacher and student models, measured by the Mean Squared Error (MSE) loss.

To this end, the full training objective is a combination of the aforementioned objectives:
\begin{equation}
	\label{eq:finalloss}
	\begin{aligned}
    \mathcal{L}=\mathcal{L}_{\text{task}}(\boldsymbol{\theta},\mathbf{z})+\lambda \mathcal{L}_{\text{reg}} (\mathbf{z}) + \frac{1}{L} \sum_{i=1}^L \mathcal{L}_{\text{MSE}}(\mathbf{h}_{\text{stu}}^i, \boldsymbol{W}_{\text{layer}} \cdot \mathbf{h}_{\text{tea}}^i) 
	\end{aligned} 
\end{equation}
where $\mathbf{h}_{\text{tea}}^i$ and $\mathbf{h}_{\text{stu}}^i$ represent the $i$-th layer hidden outputs of the teacher and student models, respectively. $\boldsymbol{W}_{\text{layer}}$ denotes the transformation matrix between $\mathbf{h}_{\text{tea}}^i$ and $\mathbf{h}_{\text{stu}}^i$, which can improve the distillation performance.

\begin{table}[!ht]
    \centering
    \setlength{\tabcolsep}{1mm}
    \caption{The WER results in the Librispeech test set varied across different sparsity targets. 
    Our results (both sparsity-aware and FLOPs-aware) are represented by ``Ours (Sparsity)" and ``Ours (FLOPs)", respectively.}
    \vspace{-2ex}
    \label{tab:main_results}
    \resizebox{1\columnwidth}{!}{
    \begin{tabular}{p{28pt}<{\centering}|p{57pt}p{20pt}<{\centering}p{20pt}<{\centering}cc}
    \toprule
        Sparsity & Methods & Test-clean & Test-other & RTF & Model Size (MB)\\
         \cmidrule (lr){1-2}\cmidrule (lr){3-4} \cmidrule (lr){5-6}
     0\% & Conformer & 2.86 & 6.10 & 0.193 & 75.5  \\
     \hline
     \hline
    \multirow{5}{*}{20\%} 
    & OMP~\cite{lai2021parp} & 3.12 & 6.41 & 0.174 (-9.7\%) & 61.9 (-18.1\%) \\
    & PARP~\cite{lai2021parp} & 3.12 & 6.24 & 0.175 (-9.1\%) & 61.9 (-18.1\%) \\
    & SVD~\cite{povey2018semi} & 4.23 & 8.59 & 0.180 (-6.4\%) & 60.3 (-20.2\%) \\
    & {\cellcolor[rgb]{0.925,0.957,1}}\textbf{Ours} (Sparsity) & {\cellcolor[rgb]{0.925,0.957,1}}2.96 & {\cellcolor[rgb]{0.925,0.957,1}}\textbf{6.09} & {\cellcolor[rgb]{0.925,0.957,1}}\textbf{0.171 (-11.3\%)} & {\cellcolor[rgb]{0.925,0.957,1}}\textbf{60.2 (-20.3\%)} \\
    & {\cellcolor[rgb]{0.925,0.957,1}}\textbf{Ours} (FLOPs) & {\cellcolor[rgb]{0.925,0.957,1}}\textbf{2.88} & {\cellcolor[rgb]{0.925,0.957,1}}6.14 & {\cellcolor[rgb]{0.925,0.957,1}}0.174 (-9.9\%) & {\cellcolor[rgb]{0.925,0.957,1}}61.6 (-18.4\%) \\
    \hline
    \hline
    \multirow{5}{*}{30\%} & OMP~\cite{lai2021parp} & 3.28 & 6.99 & 0.164 (-14.9\%) & 53.8 (-28.7\%)  \\
    & PARP~\cite{lai2021parp} & 3.31 & 6.83 & 0.165 (-14.2\%) & 53.8 (-28.7\%) \\
    & SVD~\cite{povey2018semi} & 4.70 & 9.69 & 0.171 (-11.2\%) & 52.7 (-30.2\%) \\
    & {\cellcolor[rgb]{0.925,0.957,1}}\textbf{Ours} (Sparsity) & {\cellcolor[rgb]{0.925,0.957,1}}\textbf{3.05} & {\cellcolor[rgb]{0.925,0.957,1}}\textbf{6.29} & {\cellcolor[rgb]{0.925,0.957,1}}0.164 (-14.8\%) & {\cellcolor[rgb]{0.925,0.957,1}}52.1 (-31.0\%) \\
    & {\cellcolor[rgb]{0.925,0.957,1}}\textbf{Ours} (FLOPs) & {\cellcolor[rgb]{0.925,0.957,1}}3.08 & {\cellcolor[rgb]{0.925,0.957,1}}6.41 & {\cellcolor[rgb]{0.925,0.957,1}}\textbf{0.161 (-16.2\%)} & {\cellcolor[rgb]{0.925,0.957,1}}\textbf{47.3 (-37.4\%)}  \\
     \hline
     \hline
    \multirow{5}{*}{40\%} & OMP~\cite{lai2021parp} & 3.56 & 7.55 & 0.152 (-21.0\%) & 45.8 (-39.3\%) \\
    & PARP~\cite{lai2021parp} & 3.54 & 7.39 & 0.153 (-20.8\%) & 45.8 (-39.3\%)   \\
    & SVD~\cite{povey2018semi} & 4.75 & 9.76 & 0.163 (-15.6\%) & 45.2 (-40.2\%) \\
    & {\cellcolor[rgb]{0.925,0.957,1}}\textbf{Ours} (Sparsity) & {\cellcolor[rgb]{0.925,0.957,1}}\textbf{3.12} & {\cellcolor[rgb]{0.925,0.957,1}}\textbf{6.61} & {\cellcolor[rgb]{0.925,0.957,1}}\textbf{0.152 (-21.1\%)} & {\cellcolor[rgb]{0.925,0.957,1}}44.1 (-41.6\%) \\
    & {\cellcolor[rgb]{0.925,0.957,1}}\textbf{Ours} (FLOPs) & {\cellcolor[rgb]{0.925,0.957,1}}3.15 & {\cellcolor[rgb]{0.925,0.957,1}}6.88 & {\cellcolor[rgb]{0.925,0.957,1}}\textbf{0.152 (-21.1\%)} & {\cellcolor[rgb]{0.925,0.957,1}}\textbf{40.0 (-47.1\%)} \\
     \hline
     \hline
    \multirow{5}{*}{50\%} 
    & OMP~\cite{lai2021parp} & 3.57 & 8.02 & 0.145 (-24.8\%) & 37.8 (-50.0\%)  \\
    & PARP~\cite{lai2021parp} & 3.56 & 7.83 & 0.147 (-23.7\%) & 37.8 (-50.0\%) \\
    & SVD~\cite{povey2018semi} & 4.78 & 9.84 & 0.151 (-21.6\%) & 37.7 (-50.1\%)\\
    & {\cellcolor[rgb]{0.925,0.957,1}}\textbf{Ours} (Sparsity) & {\cellcolor[rgb]{0.925,0.957,1}}\textbf{3.27} & {\cellcolor[rgb]{0.925,0.957,1}}\textbf{6.89} & {\cellcolor[rgb]{0.925,0.957,1}}0.144 (-25.0\%) & {\cellcolor[rgb]{0.925,0.957,1}}37.1 (-50.9\%) \\
    & {\cellcolor[rgb]{0.925,0.957,1}}\textbf{Ours} (FLOPs) & {\cellcolor[rgb]{0.925,0.957,1}}3.29 & {\cellcolor[rgb]{0.925,0.957,1}}7.17 & {\cellcolor[rgb]{0.925,0.957,1}}\textbf{0.139 (-28.1\%)} & {\cellcolor[rgb]{0.925,0.957,1}}\textbf{36.7 (-51.4\%)} \\
    \bottomrule
    \end{tabular}
    }
\end{table}
\section{Experiments}
\subsection{Experimental Setup}
\noindent\textbf{Model and dataset}. We apply our pruning method on the Conformer-Transducer model and evaluate its performance on the standard LibriSpeech dataset~\cite{panayotov2015librispeech},  which contains 960 hours of training speech data.  Specifically, the model  consists of 18 encoder layers and 2 decoder layers. 
Each encoder layer has 512 hidden dimensions, 8 heads, 1024 FFN intermediate dimensions, and convolution with kernel size of 3. In our work, we only mask and  prune modules in encoder layers.

\noindent\textbf{Pruning}. Before pruning, we follow the training receipts in Conformer\cite{gulati2020conformer} to pre-train the model until convergence.  
Then, we set various sparsity ratios of 20\%, 30\%, 40\%, 50\%  and apply our approach on the pre-trained model under each sparsity ratio. We consider both the FLOPs sparsity ratio and model parameter size ratio as discussed in Sec.~\ref{sec:prune}. To learn the pruning masks, we utilize three AdamW~\cite{loshchilov2017decoupled} optimizers, one for the ASR task transducer loss, another for the $L_0$ loss, and the third for the Lagrangian constraint loss. The initial learning rates are set to 1e-2, -1e-2, and 3e-4, respectively. We set $\lambda$ to 1. For each sparsity ratio, 
we gradually increase the sparsity from 0 to the target value within the first 10k steps, and then keep the target value for the remaining 100k steps. All the experiments are trained on 8 NVIDIA V100 GPUs using ESPnet\footnote{https://github.com/espnet/espnet}.

\noindent\textbf{Baselines}. We compare our  approach with other widely-used compression methods: SVD~\cite{povey2018semi} and two magnitude pruning methods (One-shot Magnitude Pruning (OMP), PARD~\cite{lai2021parp}). In particular, OMP and PARD are originally implemented for unstructured pruning, which do not provide any benefits to inference latency. For a fair comparison, we implement OMP and PARD using our structured pruning granularity setting.

To evaluate the performance of the pruned models, we report the WER results on the LibriSpeech test, as well as the  encoder model size and the inference real-time factor (RTF) of the encoder part on a single-core Intel(R) Xeon(R) CPU (2.60GHz).

\subsection{Main Results}

Table \ref{tab:main_results} presents a summary of the WER results as well as RTF achieved by various methods. To begin with, we compare the performance of our pruned models with the original Conformer model (0\%). Notably, we are able to remove 20\% of the unimportant model parameters from the Conformer model without any significant loss in WER performance. This results in an 11.3\% inference acceleration. As the sparsity increases, we observe a slight drop in WER performance but with a notable improvement in inference acceleration. Interestingly, even at a high sparsity ratio of 50\%,  our pruned model still achieves better performance than the open-source Conformer\footnote{https://github.com/espnet/espnet/blob/master/egs/librispeech/asr1}, with noteworthy WER of 3.27 and 6.89 on the test-clean and test-other datasets, respectively. This translates into a 25\% reduction in RTF, demonstrating the effectiveness of our pruning approach.

Table \ref{tab:main_results} clearly demonstrates that our method outperforms all the baseline state-of-the-art structured pruning methods, achieving significantly better results at all sparsity ratios. SVD baseline exhibits poor performance on the LibriSpeech dataset, which is known to be particularly sensitive to model parameters. When compared to the magnitude-based method, our approach incurs smaller WER losses, especially on the test-other dataset, with relative WER loss percentages of 8.23\%/5.25\%, 7.91\%/11.60\%, 15.23\%/15.36\%, and 10.47\%/18.55\% at 20\%, 30\%, 40\%, and 50\% sparsity targets, respectively. Compared to  PARD, our approach achieves relative WER loss improvements of 8.30\%/2.56\%, 9.21\%/8.98\%, 14.74\%/12.82\%, and 10.01\%/15.42\% on the test-clean and test-other datasets, respectively.
Notably, as the sparsity targets increase, our method achieves a greater advantage in terms of WER loss. %further underscoring its efficacy in reducing model complexity without sacrificing performance.

\subsection{Ablation Study}
\begin{table}[t]
    \centering
    \setlength{\tabcolsep}{1mm}
    \caption{The WER results of different ablation settings on Librispeech test set.
     ``w/o layer-wise hidden masks" means we remove layer-wise hidden dimension pruning and  ``w head pruning" denotes only conducting head pruning.
    }
\vspace{-2ex}
    \label{tab:ablation}
    \resizebox{0.7\columnwidth}{!}{
    \begin{tabular}{c|lcc}
    \toprule
        Sparsity & Methods & Test-clean & Test-other \\
         \cmidrule(lr){1-2}\cmidrule(lr){3-4}
    \multirow{4}{*}{20\%} & \textbf{Ours} (Sparsity) & 2.96  & \textbf{6.09} \\
    & w/o layer-wise hidden & 2.98 & 6.24 \\
    & w head level pruning & 3.14 & 6.69 \\
    & w/o layer-wise KD & \textbf{2.95} & 6.19 \\
    \hline
    \hline
    \multirow{4}{*}{30\%} & \textbf{Ours} (Sparsity) & \textbf{3.05}  & \textbf{6.29} \\
    & w/o layer-wise hidden & 3.12 & 6.45 \\
    & w head level pruning & 3.37 & 7.03 \\
    & w/o layer-wise KD & 3.07 & 6.30 \\
     \hline
     \hline
    \multirow{4}{*}{40\%} & \textbf{Ours} (Sparsity) & \textbf{3.12}  & \textbf{6.61}  \\
    & w/o layer-wise hidden & 3.26 & 6.89\\
    & w head level pruning & 3.58 & 7.39 \\
    & w/o layer-wise KD & 3.15 & 6.73 \\
     \hline
     \hline
    \multirow{4}{*}{50\%} & \textbf{Ours} (Sparsity) & \textbf{3.27}  & \textbf{6.89} \\
    & w/o layer-wise hidden & 3.41  & 7.19 \\
    & w head level pruning & 3.72 & 7.82 \\
    & w/o layer-wise KD & 3.29 & 6.93 \\
    \bottomrule
    \end{tabular}
    }
\end{table}

We now conduct ablation studies to evaluate (1) the effectiveness of our hybrid pruning granularity and (2) the effectiveness of combining knowledge distillation with pruning. For experiment (1), we compare our current setting with head pruning (w head level pruning), which is widely-used in other NLP tasks~\cite{michel2019sixteen,voita2019analyzing}, and we compare with disabling the pruning for layer-wise hidden dimension (w/o layer-wise hidden). For experiment (2), we remove layer-wise KD for evaluation.

Table~\ref{tab:ablation} demonstrates the effectiveness of our proposed hybrid pruning approach, which enables us to selectively prune fine-grained units such as heads, Conv, FFN, and layer-wise hidden dimensions.  In contrast, conducting head pruning alone leads to a significant loss of 32.9\% in WER because crucial heads can be removed to achieve high sparsity.
Disabling layer-wise hidden pruning also slightly degrades the performance. 

Table~\ref{tab:ablation} also suggests that the proposed combination of knowledge distillation and pruning effectively improves the ASR performance. It is worth noting that layer-wise KD consistently demonstrates an absolute WER improvement of 0.04-0.13 under different sparse targets. This can be attributed to the fact that layer-wise KD facilitates the transfer of hierarchical information from the teacher model to the student model on a layer-by-layer basis,  which  encourages the pruned model to retain more semantic knowledge. 

\subsection{Analysis of Pruning Results}
\label{subsec:analysis}

\begin{figure}[t]
  \centering
    \begin{subfigure}[]{1\linewidth}
         \centering
         \includegraphics[width=\linewidth]{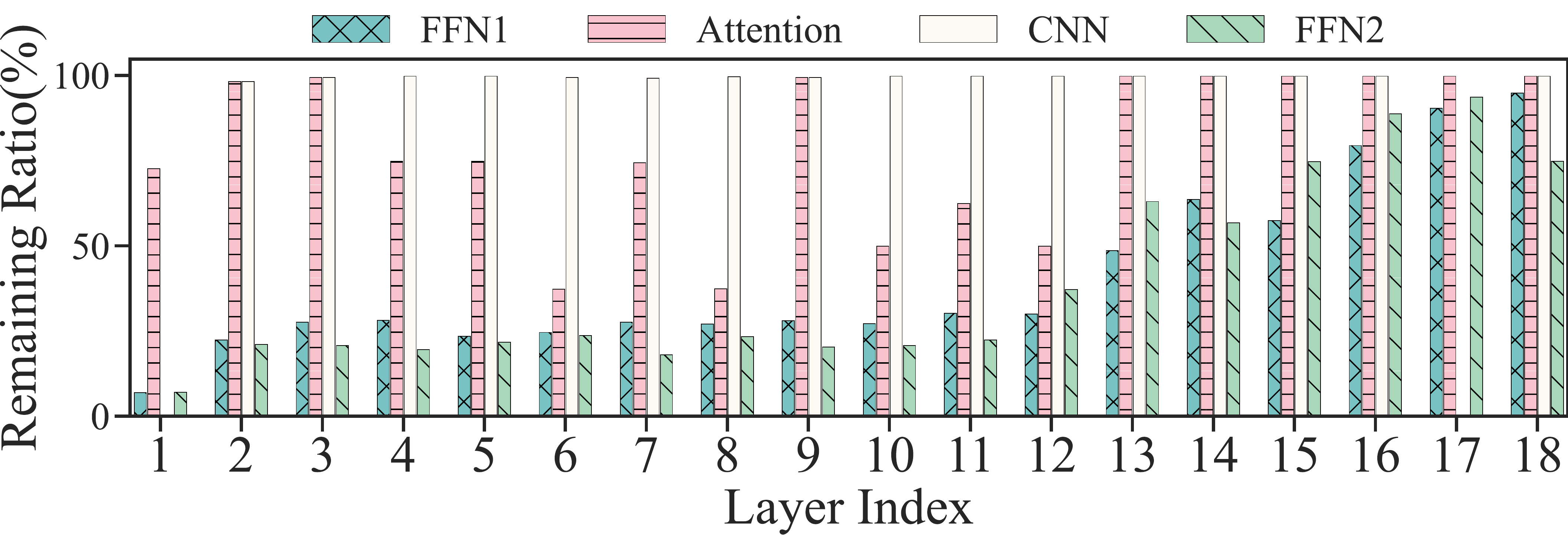}
         \caption{Sparse distribution of different modules (50\% sparsity target)}
         \label{subfig:sparse_distribution}
     \end{subfigure}
     \begin{subfigure}[]{1\linewidth}
         \centering
         \includegraphics[width=\linewidth]{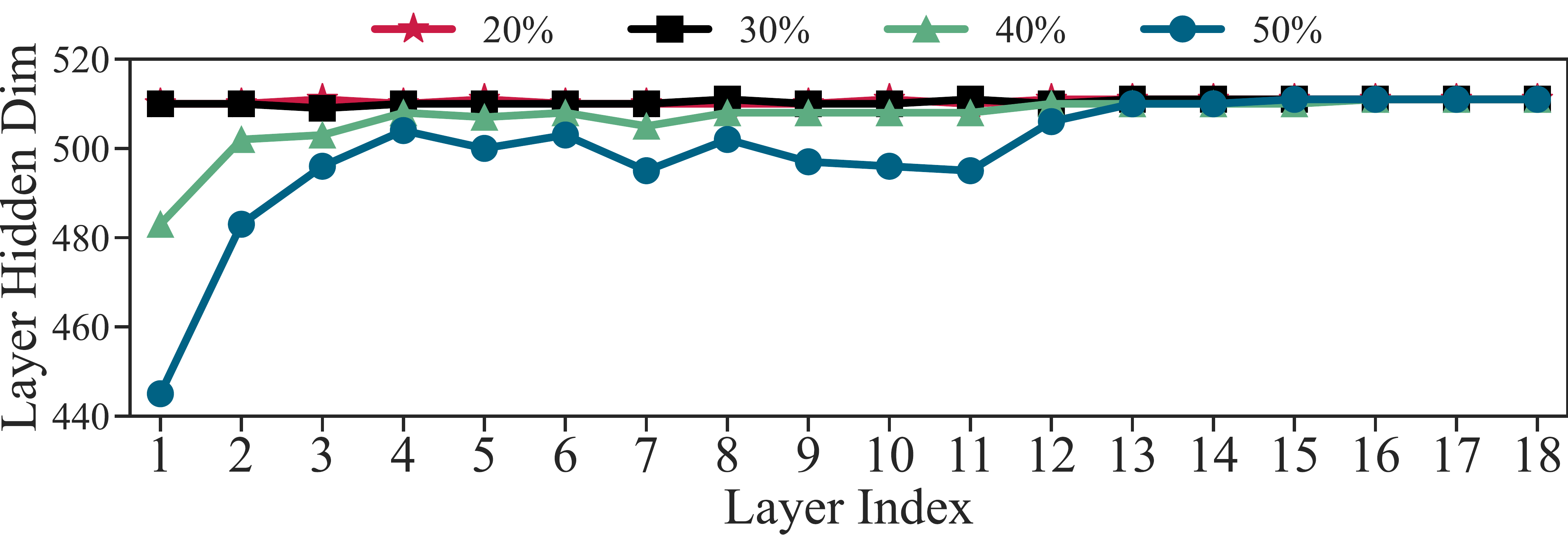}
         \caption{Layer-wise hidden dimension for the pruned model}
         \label{subfig:layer_hidden_distrubution}
     \end{subfigure}
     \vspace{-2ex}
  \caption{The distribution of the remaining ratio across different modules and layers when aiming for different sparsity targets.}
  \label{fig:layer_sparsity}
\end{figure}
Finally, we study the pruned Conformer structures produced by our approach. 
Fig.~\ref{subfig:sparse_distribution} shows the retained parameters ratios for the 50\% sparsity model. Interestingly, the Conformer model shows more redundancy in the bottom layers. Specifically,  our approach removes up to 76.7\% parameters at the bottom layers (Layer1-10), while only 6.08\%-11.9\% are removed at the top layers (Layer 16-18). The results suggest that the Conformer-Transducer model exhibits a bottom-heavy redundancy and top-sensitive pattern in the ASR task. Moreover, the model displays varied redundancy in different modules. Compared to attention and CNN modules, we prune more parameters in FFN layers. 

Fig.~\ref{subfig:layer_hidden_distrubution} shows the remaining hidden dimensions of various layers at different sparsity ratios. As the target sparsity ratio increases, more hidden dimensions are pruned. Notably, at the target sparsity of 50\%, the lower layers can prune up to 15\% of their dimensions, while deeper layers exhibit a very minimal pruning. These findings suggest that deeper layer model parameters are crucial for ASR performance, and even slight changes in their dimension sizes can have a significant impact.

\section{Conclusion}
In this paper, we propose a structured pruning algorithm that boosts the efficiency of automatic speech recognition by   pruning unimportant modules with a hybrid granularity.
 We target all levels of pruning in the Conformer model, including Attention heads, FFN layers, Conv module, and layer-wise hidden dimension. Our approach combines $L_0$ regularization and knowledge distillation to achieve optimal pruning decisions. Experiments on LibriSpeech show that our method cuts model size by 50\% and CPU inference cost by 28\%, with a minimal WER performance loss, far exceeding the baseline model.

\bibliographystyle{IEEEtran}
\bibliography{main}

\end{document}